\documentclass[sigconf]{acmart}
\usepackage[utf8]{inputenc}
\usepackage{xspace}
\usepackage{subfig}
\usepackage{multirow}

\newcommand{\method}{\textit{scMoGNN}\xspace}

\AtBeginDocument{%
  \providecommand\BibTeX{{%
    \normalfont B\kern-0.5em{\scshape i\kern-0.25em b}\kern-0.8em\TeX}}}

\copyrightyear{2022}
\acmYear{2022}
\setcopyright{acmcopyright}
\acmConference[KDD '22] {Proceedings of the 28th ACM SIGKDD Conference on Knowledge Discovery and Data Mining}{August 14--18, 2022}{Washington, DC, USA.}
\acmBooktitle{Proceedings of the 28th ACM SIGKDD Conference on Knowledge Discovery and Data Mining (KDD '22), August 14--18, 2022, Washington, DC, USA}
\acmPrice{15.00}
\acmISBN{978-1-4503-9385-0/22/08}
\acmDOI{10.1145/3534678.3539213}
\begin{document}

\title{Graph Neural Networks for \\ Multimodal Single-Cell Data Integration}

\author{Hongzhi Wen}\authornote{These authors contribute equally to this work.} 
\orcid{0000-0003-0775-8538}
\affiliation{%
  \institution{Michigan State University}
  \city{}
  \state{}
  \country{}}
\email{wenhongz@msu.edu}

\author{Jiayuan Ding}\authornotemark[1]
\affiliation{%
  \institution{Michigan State University}
  \city{}
  \state{}
  \country{}}
\email{dingjia5@msu.edu}

\author{Wei Jin}\authornotemark[1]
\affiliation{%
  \institution{Michigan State University}
  \city{}
  \state{}
  \country{}}
\email{jinwei2@msu.edu}

\author{Yiqi Wang}\authornotemark[1]
\affiliation{%
  \institution{Michigan State University}
  \city{}
  \state{}
  \country{}}
\email{wangy206@msu.edu}

\author{Yuying Xie}
\affiliation{%
  \institution{Michigan State University}
  \city{}
  \state{}
  \country{}}
\email{xyy@msu.edu}

\author{Jiliang Tang}
\affiliation{%
  \institution{Michigan State University}
  \city{}
  \state{}
  \country{}}
\email{tangjili@msu.edu}

\renewcommand{\shortauthors}{Hongzhi Wen et al.}


\begin{abstract}
Recent advances in multimodal single-cell technologies have enabled simultaneous acquisitions of multiple omics data from the same cell, providing deeper insights into cellular states and dynamics. However, it is challenging to learn the joint representations from the multimodal data, model the relationship between modalities, and, more importantly, incorporate the vast amount of single-modality datasets into the downstream analyses. To address these challenges and correspondingly facilitate multimodal single-cell data analyses, three key tasks have been introduced: \textit{Modality prediction},  \textit{Modality matching} and  \textit{Joint embedding}. In this work, we present a general Graph Neural Network framework \method{} to tackle these three tasks and show that \method{} demonstrates superior results in all three tasks compared with the state-of-the-art and conventional approaches.
Our method is an official winner in the overall ranking of \textit{Modality prediction} from NeurIPS 2021 Competition\footnote{\url{https://openproblems.bio/neurips_2021/}}, and all implementations of our methods have been integrated into DANCE package~\footnote{\url{https://github.com/OmicsML/dance}}.

\end{abstract}




\settopmatter{printacmref=true} 

\maketitle

\vskip -0.5em
\section{Introduction}

The rapid advance of single-cell technologies makes it possible to simultaneously measure multiple molecular features at multiple modalities in a cell, such as gene expressions, protein abundance and chromatin accessibility. For instance, CITE-seq (cellular indexing of transcriptomes and epitopes by sequencing)~\cite{stoeckius2017simultaneous} enables simultaneous quantification of mRNA expression and surface proteins abundance; methods like sci-CAR~\cite{cao2018joint}, Paired-seq~\cite{zhu2019ultra}, and SNARE-seq~\cite{chen2019high} enable joint profiling of mRNA expression and chromatin accessibility ( i.e. genome-wide DNA accessibility).
The joint measurements from these methods provide unprecedented multimodal data for single cells, which has given rise to valuable insights for not only the relationship between different modalities but, more importantly, a holistic understanding of the cellular system.


Despite the emergence of joint platforms, single-modality datasets are still far more prevalent. How to effectively utilize complementary information from multimodal data to investigate cellular states and dynamics and to incorporate the vast amount of single-modality data while leveraging the multimodal data pose great challenges in single-cell genomics. To address these challenges, Luecken~\citet{luecken2021a} summarized three major tasks: (1) \textit{Modality prediction} aims at predicting the features of one modality from the features of another modality~\cite{wu2021babel}; (2) \textit{Modality matching} focuses on identifying the correspondence of cells between different modalities~\cite{welch2017matcher}; and (3) \textit{Joint embedding} requires embedding the features of two modalities into the same low-dimensional space~\cite{stoeckius2017simultaneous}.
The motivation of modality prediction and modality matching is to better integrate existing single-modality datasets, while joint embedding can provide more meaningful representations of cellular states from different types of measurements. In light of these benefits, computational biologists recently organized a competition for multimodal single-cell data integration at NeurIPS 2021~\cite{luecken2021a} to benchmark these three tasks and facilitate the computational biology communities.

There is an emerging trend to leverage deep learning techniques to tackle the tasks mentioned above for multimodal single-cell data~\cite{molho2022deep}. BABEL~\cite{wu2021babel} translated between the transcriptome (mRNA) and chromatin (DNA) profiles of a single cell based on an encoder-decoder architecture;  scMM~\cite{minoura2021mixture} implemented a mixture-of-experts deep generative model for joint embedding learning and modality prediction. Cobolt~\cite{gong2021cobolt} acquired joint embedding via a variant of Multimodal Variational Autoencoder (MVAE~\cite{yao2021transcriptomic}). MOFA2~\cite{argelaguet2020mofa+} used Bayesian group factor analysis to reduce dimensions of multi-modality data and generate a low-dimensional joint representation. However, most of these approaches treat each cell as a separate input without considering possible high-order interactions among cells or different modalities. Such higher-order information can be essential for learning with high-dimensional and sparse cell features, which are common in single-cell data. Take the joint embedding task for example, the feature dimensions for GEX (mRNA) and ATAC (DNA) data are as high as 13,431 and 116,490, respectively; however, only $9.75\%$ of GEX and $2.9\%$ of ATAC features are nonzero on average over $42,492$ training samples (cells).
Furthermore, integrated measuring often requires additional processing to cells, which can lead to extra noise and drop-out in the resulting data~\cite{lee2020single, mimitou2021scalable}. Therefore, it is a desired technique that can mitigate the negative impact of such noise.

Recently, the advances in graph neural networks (GNNs)~\cite{kipf2016semi,wu2020comprehensive,battaglia2018relational,gilmer2017neural,liu2021graph} pave the way for addressing the aforementioned issues in single-cell data integration. Specifically, GNNs aggregate information from neighborhoods to update node embeddings iteratively~\cite{gilmer2017neural}. Thus, the node embedding can eventually encode high-order structural information through multiple aggregation layers. In addition, GNNs smooth the features by aggregating neighbors' embedding and also filter the eigen-values of graph Laplacian, which provides an extra denoising mechanism~\cite{ma2021unified}. Hence, by modeling the interactions between cells and their features as a graph, we can adopt GNNs to exploit the structural information and tackle the limitations of previous techniques for single-cell data integration. With the constructed graph, we can readily incorporate external knowledge (e.g., interactions between genes) into the graph to serve as additional structural information. Moreover, it enables a transductive learning paradigm with GNNs to gain additional semi-supervised signals to enhance representation learning. 

Given those advantages, we aim to design a GNN framework for multimodal data integration. While several existing works attempted to introduce graph neural networks to single cell analysis~\cite{song2021scgcn,wang2021scgnn,ciortan2022gnn,shao2021scdeepsort}, none of them tackle the challenging problem of multimodal data integration which requires handling different modalities simultaneously. Therefore, we aim to develop GNN methods for the tasks in multimodal data integration, especially for modality prediction, modality matching and joint embedding. Specifically, we propose a general framework \method{} for modeling interactions of \underline{mo}dalities and leveraging \underline{GNN}s in \underline{s}ingle-\underline{c}ell analysis\footnote{Our solution won the first place of the modality prediction task in the Multimodal Single-Cell Data Integration competition at NeurIPS 2021.}. Our framework is highly versatile: we demonstrate its use cases in the three different multimodal tasks. To the best of our knowledge, we are the first to develop a GNN framework in this emerging research topic, i.e., multimodal single-cell data integration. Our proposed framework achieves the best results in all of these three tasks on the benchmark datasets, providing a very strong baseline for follow-up research. Our contributions can be summarized as follows:
\begin{enumerate}
\item We study the problem of multimodal single-cell data integration and propose a general GNN-based framework \method{} to capture the high-order structural information between cells and modalities. 
\item The proposed general framework is highly flexible as it can be adopted in different multimodal single-cell tasks. 
\item Our framework achieves remarkable performance across tasks. It has won the first place of the modality prediction task in the Multimodal Single-Cell Data Integration competition, and currently outperforms all models for all three tasks on the leaderboard\footnote{\url{https://eval.ai/web/challenges/challenge-page/1111/leaderboard/2860}}. All of our results are based on publicly available data and are reproducible.
\end{enumerate}


\section{Related Work}
In this section, we briefly introduce works related to our work including GNNs on single-modality data and multimodal data integration.

\noindent\textbf{GNNs on Single-Modality Data.} 
Graphs occur as a natural representation of 
single-cell data both as feature-centric (RNAs, DNAs, or proteins) and cell-centric. 
Thus, a few recent works have applied GNNs to the single-cell data. 
~\citet{song2021scgcn} propose scGCN model for knowledge transfer in single-cell omics (mRNA or DNA) based on Graph Convolutional Networks~\cite{kipf2016semi}. 
scGNN~\cite{wang2021scgnn} formulates and aggregates cell-cell relationships with Graph Neural Networks for missing-data imputation and cell clustering using single-cell RNA sequencing (scRNA-seq) data. scDeepSort~\cite{shao2021scdeepsort} is a pre-trained cell-type annotation tool for scRNA-seq data that utilizes a deep learning model with a weighted GNN. Similar to our proposed model, scDeepSort also relies on feature-cell graphs. However, it does not incorporate any prior knowledge into GNNs. 
 Using spatial transcriptomics (mRNA) data, DSTG~\cite{song2021dstg} utilizes semi-supervised GCN to deconvolute the relative abundance of different cell types at each spatial spot. Despite its success on single-modality data, there are few efforts on applying GNNs to multimodal single-cell data. 
 
\noindent\textbf{Multimodal Data Integration.}
Most of the prior works in multimodal data integration can be divided into 1) matrix factorization ~\cite{duren2018integrative, stein2018enter, jin2020scai} or statistical based methods ~\cite{stuart2019comprehensive, shen2009integrative, welch2017matcher} and 
2) autoencoder based methods ~\cite{wu2021babel, gong2021cobolt}. Specifically, BABEl~\cite{wu2021babel} leverages autoencoder frameworks with two encoders and two decoders to take only one of these modalities and infer the other by constructing reconstruction loss and cross-modality loss. Cobolt\cite{gong2021cobolt} acquires joint embedding via a variant of Multimodal Variational Autoencoder (MVAE\cite{yao2021transcriptomic}). Unlike our proposed models, these aforementioned methods are unable to incorporate high-order interactions among cells or different modalities. 
To the best of our knowledge, we are the first to apply GNNs in the field of multimodal single-cell data integration and build a GNNs-based general framework to broadly work on these three key tasks 
from NeurIPS 2021 Competition\footnote{\url{https://openproblems.bio/neurips_2021/}}. Our framework officially won first place in the overall ranking of the modality prediction task. After the competition, we extended our framework to the other two tasks and achieved superior performance compared with the top winning methods.

\section{Problem Statement}


Before we present the problem statement, we first introduce the notations used in this paper. There are three modalities spanning through each task. They are GEX as mRNA data, ATAC as DNA data and ADT as protein data. Each modality is initially represented by a matrix $\mathbf{M} \in \mathbb{R}^{N \times k}$ where $N$ indicates the number of cells, and $k$ denotes the feature dimension for each cell. In our work, we later construct a bipartite graph $\mathcal{G}=(\mathcal{U}, \mathcal{V},\mathcal{E})$ based on each modality $\mathbf{M}$, where $\mathcal{U}$ is the set of $N$ cell nodes $\{u_1, u_2, ..., u_N\}$ and $\mathcal{V}$ is the set of $k$ feature nodes $\{v_1, v_2, ..., v_k\}$.

With the aforementioned notations, the problem of learning GNNs for single-cell data integration is formally defined as,
\vskip 0.5em
\noindent\textit{Given a modality $\mathbf{M} \in \mathbb{R}^{N \times k}$, we aim at learning a mapping function  $f_\theta$ which maps $\mathbf{M}$ to the space of downstream tasks.}
\vskip 0.5em

In the following, we formally define these three key tasks of single-cell data integration: modality prediction, modality matching and joint embedding. We will also define the corresponding evaluation metrics for each task. Note that these metrics are also adopted by the competition to decide the top winners. 

\subsection{Task 1: Modality Prediction}
In this task, given one modality (like GEX), the goal is to predict the other (like ATAC) for all feature values in each cell. It can be formally defined as,

\textit{Given a source modality $\mathbf{M_1} \in \mathbb{R}^{N \times k1}$, the goal is to predict a target  modality  $\mathbf{M_2} \in \mathbb{R}^{N \times k2}$ via learning a mapping function $f_\theta$ parameterized by $\theta$ such that 
$\mathbf{M_2} = f_\theta(\mathbf{M_1})$.}

Possible modality pairs of ($\mathbf{M_1}$, $\mathbf{M_2}$) are (GEX, ATAC), (ATAC, GEX), (GEX, ADT) and (ADT, GEX), which correspond to four sub-tasks in Task 1. Root Mean Square Error\footnote{\url{https://en.wikipedia.org/wiki/Root-mean-square_deviation}} is used to quantify performance between observed and predicted feature values.

\subsection{Task 2: Modality Matching}
The goal of this task is to identify the correspondence between two single-cell profiles and provide the probability distribution of these predictions. It can be formally defined as,

\textit{Given modality $\mathbf{M_1} \in \mathbb{R}^{N \times k1}$ and modality $\mathbf{M_2} \in \mathbb{R}^{N \times k2}$, we aim to learn two mapping functions $f_{\theta_1}$ parameterized by $\theta_1$ and $f_{\theta_2}$ parameterized by $\theta_2$ to map them into the same space such that
\begin{equation}
 \mathbf{S} = g( f_{\theta_1}(\mathbf{M_1}), f_{\theta_2}(\mathbf{M_2}) )
\end{equation}
where $g$ is a score function to calculate probability distribution of correspondence predictions. $\mathbf{S} \in \mathbb{R}^{N \times N}$ is an output score matrix with each row summing to 1. $\mathbf{S_{ij}}$ is the correspondence probability between $i$-th cell from modality $\mathbf{M_1}$ and $j$-th cell from modality $\mathbf{M_2}$.}

Possible modality pairs of ($\mathbf{M_1}$, $\mathbf{M_2}$) are (GEX, ATAC), (ATAC, GEX), (GEX, ADT) and (ADT, GEX), which correspond to four sub-tasks in Task 2.
The sum of weights in the correct correspondences of $\mathbf{S}$ is used as final score to quantify prediction performance using $score = \sum_{i=1}^N\sum_{j=1}^N\mathbf{S}_{i,j}$ if $i=j$.

\subsection{Task 3: Joint Embedding}
In this task, the goal is to learn an embedded representation that leverages the information of two modalities. The quality of the embedding will be evaluated using a variety of criteria generated from expert annotation.
It can be formally defined as,

\textit{Given modality $\mathbf{M_1} \in \mathbb{R}^{N \times k1}$ and modality $\mathbf{M_2} \in \mathbb{R}^{N \times k2}$, 
we aim to learn three mapping functions $f_{\theta_1}$, $f_{\theta_2}$ and $f_{\theta_3}$ parameterized by $\theta_1$, $\theta_2$ and $\theta_3$ accordingly to project them into downstream tasks, 
\begin{equation}
 \mathbf{H} = f_{\theta_3}\big(CONCAT( f_{\theta_1}(\mathbf{M_1}), f_{\theta_2}(\mathbf{M_2}) )\big)
\end{equation}
where $f_{\theta_1}(\mathbf{M_1}) \in \mathbb{R}^{N \times k1^\prime}$ and $f_{\theta_2}(\mathbf{M_2}) \in \mathbb{R}^{N \times k2^\prime}$ correspond to new representations learned from modality $\mathbf{M_1}$ and $\mathbf{M_2}$ separately, $\mathbf{H} \in \mathbb{R}^{N \times k3}$ is a final output embedding learned through $f_{\theta_3}$ on concatenation of $f_{\theta_1}(\mathbf{M_1})$  and $f_{\theta_2}(\mathbf{M_2})$.}

$\mathbf{H}$ will be measured using six different metrics broken into two classes: biology conservation and batch removal.
Biology conservation metrics include ``NMI cluster/label'', ``Cell type ASW'', ``Cell cycle conservation'' and ``Trajectory conservation'' which aim to measure how well an embedding conserves expert-annotated biology. Batch removal metrics include ``Batch ASW'' and ``Graph connectivity'' which are to evaluate how well an embedding removes batch variation. Please refer to the Appendix~\ref{sec:metric_task3} for more details about these metrics description.
Possible modality pairs of ($\mathbf{M_1}$, $\mathbf{M_2}$) are (GEX, ATAC) and (ADT, GEX), which correspond to two sub-tasks in Task 3.

In this work, we instantiate $f_\theta$ as a graph neural network model by first constructing a bipartite graph $\mathcal{G}=(\mathcal{U}, \mathcal{V},\mathcal{E})$ based on modality $\mathbf{M}$, and then learning better cell node representation through message passing on graphs.

\section{Method}
\begin{figure*}[t]%
     \centering
     \subfloat[Graph Construction]{\label{fig:graph_con}{\includegraphics[width=0.5\linewidth]{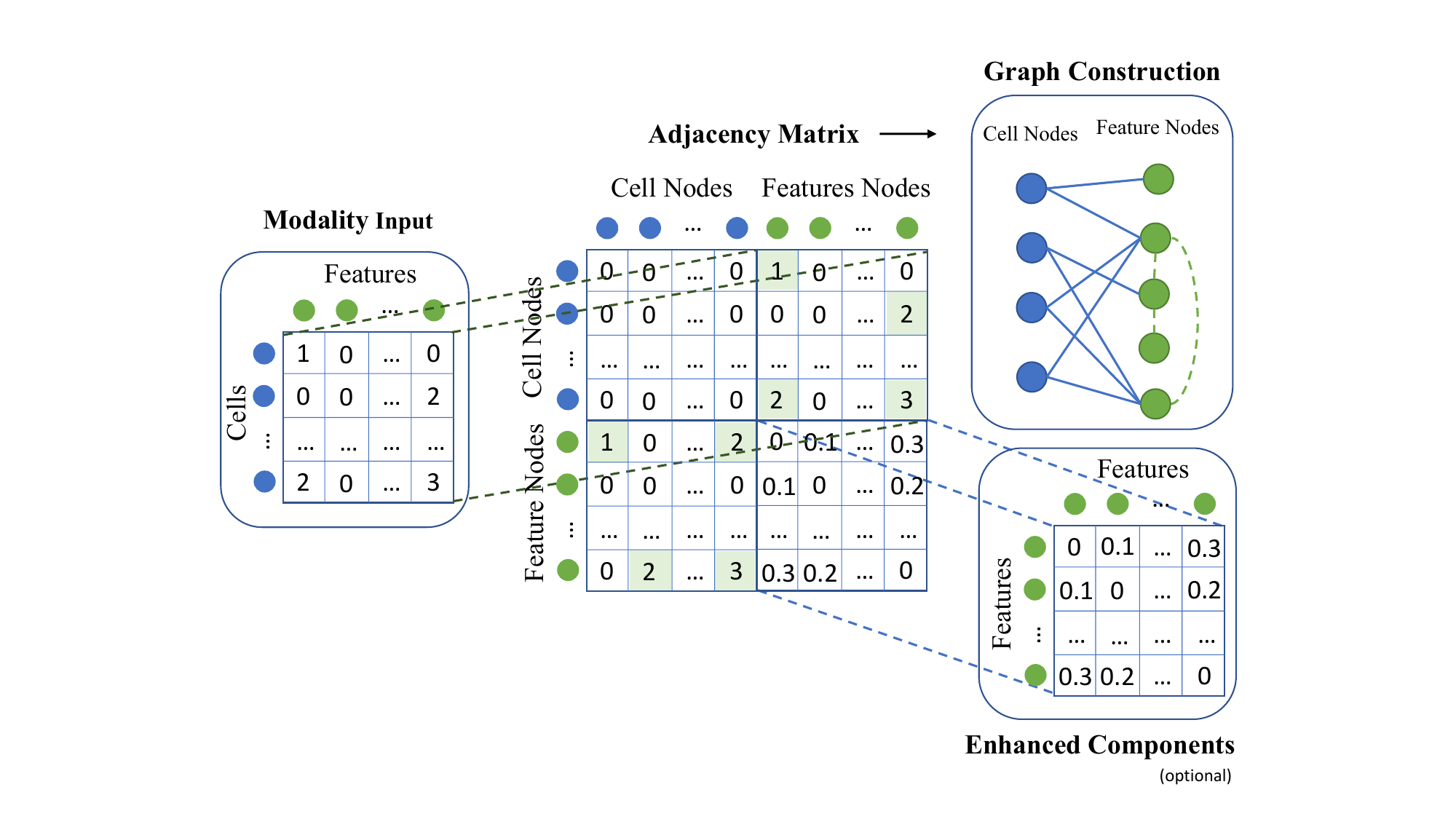} }}%
      \subfloat[Cell-feature Graph Convolution with Task-Specific Head]{\label{fig:gcn}{\includegraphics[width=0.5\linewidth]{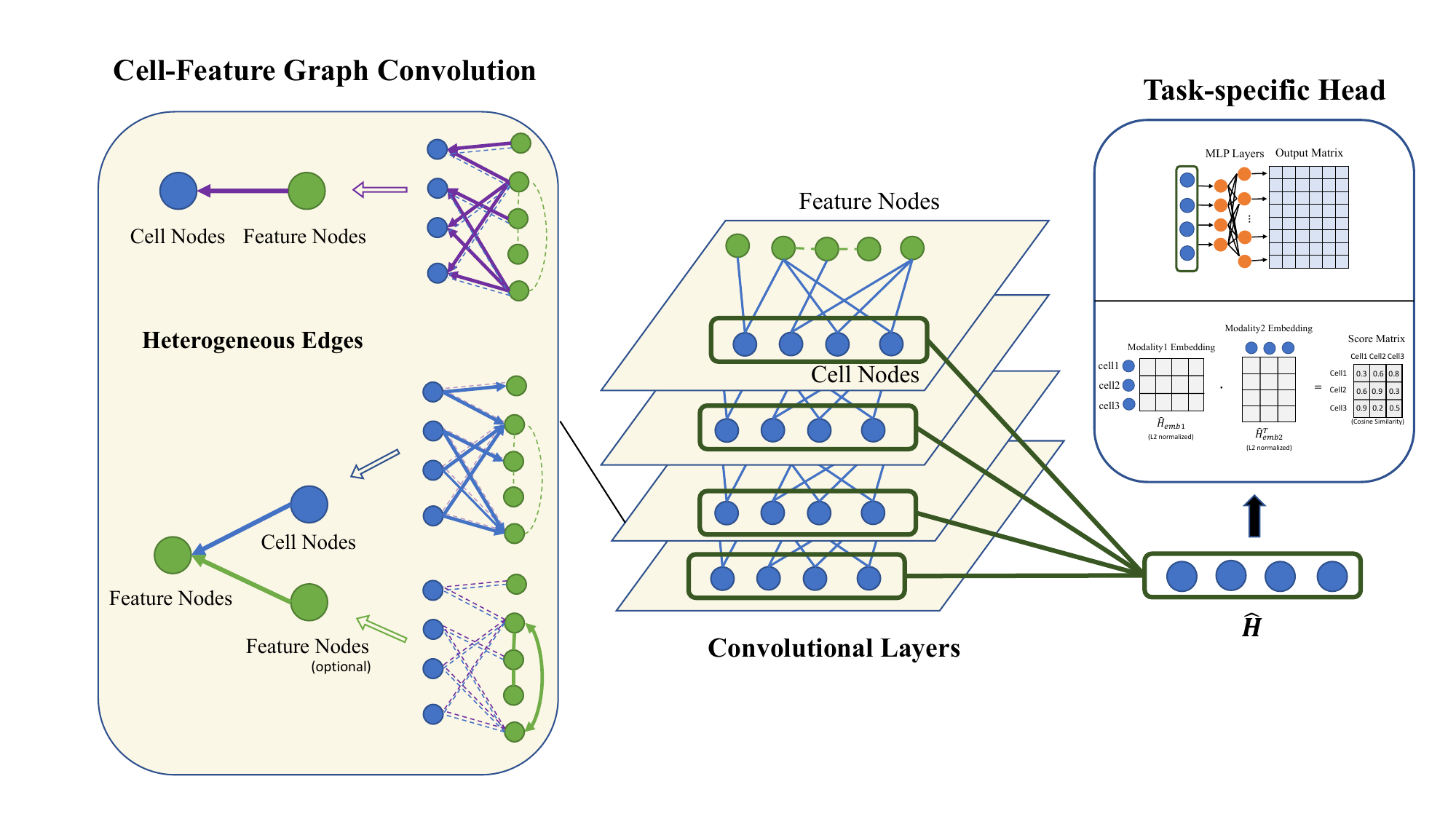} }}%
    \qquad
    \vskip -0.5em
\caption{An overview of \method{}. We first construct the cell-feature graph from a given modality and then perform cell-feature graph convolution to obtain latent embeddings of cells, which are sent to a task-specific head to perform the downstream task.}
\label{fig:framework}
\vskip -1em
\end{figure*}

In this section, we first introduce the proposed general framework \method{} for multimodal data integration. Then we introduce how to adapt \method{} to advance three tasks: modality prediction, modality matching and joint embedding. An illustration of our framework is shown in Figure~\ref{fig:framework}. Specifically, our framework can be divided into three stages: graph construction, cell-feature graph convolution, and task-specific head. 



\subsection{A General GNN Framework}
To develop a GNN-based framework for single cell data integration, we are essentially faced with the following challenges: (1) how to construct the graph for cells and its features (or modalities); (2) how to effectively extract meaningful patterns from the graph; and (3) how to adapt the framework to different multimodal  tasks.




\subsubsection{Graph Construction} With the single-cell data, our first step is to construct a cell-feature graph that GNNs can be applied to. We construct a cell-feature bipartite graph where the cells and their biological features (e.g. GEX, ADT or ATAC features) are treated as different nodes, which we term as  cell nodes and  feature nodes, respectively. A cell node is connected with the feature nodes that represent its features. With such graph structure, the advantage is that cell nodes can propagate features to their neighboring feature nodes, and vice versa. 

Formally, we denote the bipartite graph as $\mathcal{G}=(\mathcal{U}, \mathcal{V},\mathcal{E})$. In this graph, $\mathcal{U}$ is the set of $N$ cell nodes $\{u_1, u_2, ..., u_N\}$ and $\mathcal{V}$ is the set of $k$ feature nodes $\{v_1, v_2, ..., v_k\}$, where each feature node refers to one feature dimension of the input data. $\mathcal{E} \subseteq \mathcal{U} \times \mathcal{V}\ $ represents the set of edges between $\mathcal{U}$ and $\mathcal{V}$, which describe the relations between cells and features. The graph can be denoted as a weighted adjacency matrix
\begin{equation}
  \mathbf{A} = \begin{pmatrix} \mathbf{O} & \mathbf{M} \\ \mathbf{M}^T & \mathbf{O}  \end{pmatrix} \in  \mathbb{R}^{(N+k) \times (N+k)},
\end{equation}
where ${\bf O}$ is a matrix filled with constant $0$ and $\mathbf{M} \in \mathbb{R}^{N \times k}$ is the input feature matrix of cells. ${\bf M}$ can be also viewed as the features of one modality such as GEX. Note that ${\bf A}$ is a bipartite graph where two nodes within the same set (either feature nodes or cell nodes) are not adjacent. Based on the aforementioned process of graph construction, we can adjust it for specific tasks, e.g., incorporating a prior knowledge graph of genes, which can change the bipartite characteristic of ${\bf A}$. 

Furthermore, since GNNs are mostly dealing with attributed graphs, we need to assign initial embeddings for feature and cell nodes. Specifically, we use ${\mathbf{X}_{\text{cell}}}$ and $\mathbf{X}_{\text{feat}}$ to denote the initial embeddings for cell and feature nodes, respectively. We have ${\mathbf{X}_{\text{cell}}} \in \mathbb{R}^{N \times d'}$ and $\mathbf{X}_{\text{feat}} \in \mathbb{R}^{k \times d''}$ where $d'$ and $d''$ are determined by the task-specific settings. As an illustrative example, the initial embeddings of feature nodes $\{v_1,  ..., v_k\}$ could be the one-hot index of each feature dimension; thus, ${\mathbf{X}_{\text{feat}}} \in \mathbb{R}^{k \times k}$ is an identity matrix, i.e., $\mathbf{X}_\text{feat} = \mathbf{I}_{k}$.  Meanwhile, we do not have any prior knowledge for each cell, thus, ${\mathbf{X}_{\text{cell}}} = \mathbf{O}_{N \times 1}$. Together with the node features, the constructed cell-feature graph can be denoted as $\mathcal{G}=({\bf A}, \mathbf{X}_\text{feat}, \mathbf{X}_\text{cell})$. 

\subsubsection{Cell-Feature Graph Convolution} After we obtain the constructed cell-feature graph, the next challenge is how to extract high-order structural information from the graph. In this work, we utilize GNNs to effectively learn cell/feature representations over the constructed graph. Note that there are two types of nodes in the graph and we need to deal with them differently. For the ease of illustration,  we first only consider one type of nodes (e.g., feature nodes $\mathcal{V}$), and later we extend it to the whole graph. Typically, GNNs follow the message-passing paradigm~\cite{gilmer2017neural} and in each layer of GNNs, the embedding of each node is updated according to the messages passed from its neighbors. Let $\mathbf{H}^l = \{\mathbf{h}^l_1, ..., \mathbf{h}^l_N\}$, $\mathbf{h}^l_i \in \mathbb{R}^d_l$ be the input node embeddings in the $l$-th layer, where $\mathbf{h}^l_i$ corresponds to node $v_i$. Hence, the output embeddings of the $l$-th layer can be expressed as follows:
\begin{equation}
    \mathbf{h}^{l+1}_i = \operatorname{Update} (\mathbf{h}^{l}_i, \operatorname{Agg} (\mathbf{h}^{l}_j | j \in \mathcal{N}_i)),
\end{equation}
 where $\mathcal{N}_i$ is the set of first-order neighbors of node $v_i$, $\operatorname{Agg}(\cdot)$ indicates an aggregation function on neighbor nodes' embedding, and $\operatorname{Update}(\cdot)$ is an update function that generates a new node embedding vector from the previous one and aggregation results from neighbors. Notably, for the input node embedding in the first layer, we have
 \begin{equation}
 \mathbf{H}^1 = \operatorname{\sigma}(\mathbf{X}_\text{feat} \mathbf{W}_\text{feat} + \mathbf{b}_\text{feat}), \mathbf{W}_\text{feat} \in \mathbb{R}^{d' \times d}   
\end{equation}
where ${\bf W}_\text{feat}$ is a transformation matrix, $d$ is the dimension of the hidden space and $\operatorname{\sigma}$ is an activation function.
 
Though there exist a number of different GNN models, in this work, we focus on the most representative one, Graph Convolution Network (GCN)~\cite{kipf2016semi}. Note that it is straightforward to extend the proposed framework to other GNN models. Considering that we have two different types of nodes in the graph (i.e., cells and features nodes), we make some modifications on the vanilla GCN to deal with different types of nodes/edges. We name it as \textit{cell-feature graph convolution}, where we separately perform the aggregation function on different types of edges to capture interactions between cell and feature nodes. Moreover, we use different parameters for aggregation on different edges, thus allowing the embedding of each node type to have very different distributions. Specifically, from a message passing perspective, the operation in a  cell-feature graph convolution layer can be expressed as two steps, i.e., aggregation and updating. In order to generate a message ${\bf m}$ for different types of nodes, there are at least two basic aggregation functions, one is:
 \begin{equation} \label{eq: uv}
    \mathbf{m}^{i, l}_{\mathcal{U} \to \mathcal{V}} = \operatorname{\sigma}(\mathbf{b}^{l}_{\mathcal{U} \to \mathcal{V}} + \sum_{j \in \mathcal{N}_i, v_i\in \mathcal{V}} \frac{e_{ji}}{c_{ji}} \mathbf{h}_j^{l} \mathbf{W}^{l}_{\mathcal{U} \to \mathcal{V}}) 
\end{equation}
where $i$ corresponds to node $v_i \in \mathcal{V}$, $j$ corresponds to node $u_j \in \mathcal{U}$;  $e_{ji}$ denotes the edge weight between $v_j$ and $u_i$, $\mathbf{W}^{l}_{\mathcal{U} \to \mathcal{V}}$ and $\mathbf{b}^{l}_{\mathcal{U} \to \mathcal{V}}$ are trainable parameters, $\operatorname{\sigma}(\cdot)$ is an activation function such as ReLU, and $c_{ji}$ is a normalization term defined as follows:
\begin{equation} \label{eq:norm}
    c_{ji} = \sqrt{\sum_{k\in\mathcal{N}_j}e_{jk}}\sqrt{\sum_{k\in\mathcal{N}_i}e_{ki}}
\end{equation}
Obviously, Eq.~\eqref{eq: uv} is the aggregation function from cell nodes $\mathcal{U}$ to feature nodes $\mathcal{V}$. Thus the other aggregation function from $\mathcal{V}$ to $\mathcal{U}$ can be written as:
 \begin{equation} \label{eq: vu}
    \mathbf{m}^{i, l}_{\mathcal{V} \to \mathcal{U}} = \operatorname{\sigma}(\mathbf{b}^{l}_{\mathcal{V} \to \mathcal{U}} + \sum_{j \in \mathcal{N}_i, u_i\in \mathcal{U}} \frac{e_{ji}}{c_{ji}} \mathbf{h}_j^{l} \mathbf{W}^{l}_{\mathcal{V} \to \mathcal{U}})
\end{equation}
The transformation matrices $\mathbf{W}^{l}_{\mathcal{U} \to \mathcal{V}}$ and $\mathbf{W}^{l}_{\mathcal{V} \to \mathcal{U}}$ project the node embeddings from one hidden space to another vice versa. After generating the messages from neighborhoods, we then update the embedding for nodes in $\mathcal{V}$ and $\mathcal{U}$ accordingly:
\begin{equation} \label{eq:v}
    \mathbf{h}_i^{l+1} = \mathbf{h}_i^{l} + \mathbf{m}^{i, l}_{\mathcal{U} \to \mathcal{V}}, \quad     \mathbf{h}_j^{l+1} = \mathbf{h}_j^{l} + \mathbf{m}^{j, l}_{\mathcal{V} \to \mathcal{U}},
\end{equation}
where $v_i \in \mathcal{V}$ and $u_j \in \mathcal{U}$. In Eq.~\eqref{eq:v}, we adopt a simple residual mechanism in order to enhance self information. As mentioned earlier, we can have more than two types of edges depending on the downstream task and the graph structure can be much more complex than a bipartite graph. Despite such complexity, our proposed framework and cell-feature graph convolution have the capacity to handle more types of edges/nodes. We will introduce these details in Section~\ref{sec:model_spec}.
 
\subsubsection{Task-specific Head} After we learn node embeddings for feature and cell nodes, we need to project the embedding to the space of the specific downstream task. Hence, we design a task-specific head, which  depends on the downstream task. Specifically, we first take the node embeddings of cell nodes from each convolution layer, aggregate them and project them into the space of downstream task $\mathbf{\hat{Y}}$:
\begin{align}
    \mathbf{\hat{Y}} = \operatorname{Head}\left( \operatorname{Readout_{\theta}}(\mathbf{H}_\mathcal{U}^1, ..., \mathbf{H}_\mathcal{U}^L) \right)
\end{align}
where ${\bf H}^i_\mathcal{U}$ refers to embeddings of all cell nodes in $i$-th layer,  $\operatorname{Readout(\cdot)_\theta}$ is a trainable aggregation function, $\operatorname{Head(\cdot)}$ is a linear transformation that projects the latent embedding to the downstream task space. With the obtained output, we can then optimize the framework through minimizing the task-specific loss functions. In the following subsection, we give the details of training the general framework for different tasks.


\subsection{Model Specifications}\label{sec:model_spec}
With the proposed general framework \method{}, we are now able to perform different tasks by adjusting some of the components. In this subsection, we show how \method{} is applied in the three important tasks in single-cell data integration: modality prediction, modality matching and joint embedding.

\subsubsection{Modality Prediction} In the modality prediction task, our objective is to translate the data from one modality to another. 
Given the flexibility of graph construction and GNNs in our framework, we can readily incorporate external knowledge into our method. In this task, we adjust the graph construction to include such domain knowledge to enhance the feature information. Specifically, in GEX-to-ADT and GEX-to-ATAC subtasks, we introduce hallmark gene sets\cite{liberzon2015molecular} (i.e., pathway data) from the Molecular Signatures Database (MSigDB)\cite{subramanian2005gene}. The pathway data describe the correlations between gene features (i.e., features of GEX data) and the dataset consists of 50 so-called gene sets. In each gene set, a group of correlated genes are collected. However, there is no numerical information in these gene sets to help us quantify the relations between those genes, resulting in homogeneous relations. Intuitively, we can construct a fully-connected inter-gene graph for each gene set, and incorporate those edges into our original graph $\mathcal{G}$. Hence, the new adjacency matrix for $\mathcal{G}$ is:
\begin{equation}
  \mathbf{A} = \begin{pmatrix} \mathbf{O} & \mathbf{M} \\ \mathbf{M}^T & \mathbf{P} \end{pmatrix}
\end{equation}
where $\mathbf{P} \in \mathbb{R}^{k \times k}$ is a symmetric matrix, which refers to the links between gene features,  generated from gene sets data. Furthermore, we manually add some quantitative information, i.e.,  cosine similarity between gene features based on their distributions in  GEX input data.


Due to the existence of extra type of edges in the graph (i.e., edges among feature nodes), we need to make corresponding adjustment on our cell-feature graph convolution. Taking an arbitrary feature node $v_{i}$ as example, we have $\mathcal{N}_{i} = \mathcal{N}_{i}^\text{u} \cup \mathcal{N}_{i}^\text{v}$, where $\mathcal{N}_{i}$ denotes the set of neighbors of node $v_{i}$, $\mathcal{N}_{i}^\text{u} \subseteq \mathcal{U}$ is the set of cell node neighbors of $v_{i}$, $\mathcal{N}_{i}^\text{v} \subseteq \mathcal{V}$ is the set of feature node neighbors of $v_{i}$. Since cell and feature nodes have very different characteristics, when we aggregate the embedding from $\mathcal{N}_{i}^\text{u}$ and $\mathcal{N}_{i}^\text{v}$ respectively, we expect to get very different results. Thus, when we update the embedding of center node, we have different strategies to combine messages from different channels. As a starting point, we decide to enable a scalar weight. Formally speaking, similar to Eq.~\eqref{eq: uv} and Eq.~\eqref{eq: vu}, we first generate two messages $\mathbf{m}^{i, l}_{\mathcal{U} \to \mathcal{V}}$ and $\mathbf{m}^{i, l}_{\mathcal{V} \to \mathcal{V}}$ for each node $v_i \in \mathcal{V}$, then we update the node embedding of $v_i$ following the formulation below:
 \begin{equation} \label{eq:a}
    \mathbf{h}_i^{l+1} = \mathbf{h}_i^{l} + \alpha \cdot \mathbf{m}^{i, l}_{\mathcal{V} \to \mathcal{V}} + (1-\alpha) \cdot \mathbf{m}^{i, l}_{\mathcal{U} \to \mathcal{V}}
\end{equation}
where $\alpha$ is either a hyper-parameter or a learnable scaler to control the ratio between inter-feature aggregation and cell-feature aggregation.


Next we elaborate on the modality prediction head and loss function for the task. The head structure for modality prediction is relatively simple. Note that we deliberately keep the same hidden dimension throughout the cell-feature graph convolution; thus we can simply use a trainable weight vector $\mathbf{w}$ to sum up cell node embeddings from different layers, as follows: 
\begin{equation} \label{eq:ro}
    \mathbf{\hat{H}} = \sum_{i = 1}^{L} \mathbf{w}_i \cdot \mathbf{H}^i_\mathcal{U}
\end{equation}
where $\mathbf{\hat{H}}, \mathbf{H}^i_\mathcal{U} \in \mathbb{R}^{N \times d}$, and $d$ is the dimension of hidden layer. After that, a simple fully connected layer is performed to transform it to the target space:
\begin{equation}
    \mathbf{\hat{Y}} = \mathbf{\hat{H}} \mathbf{W} + \mathbf{b}.
\end{equation}
A rooted mean squared error (RMSE) loss is then calculated as:
\begin{equation}
    \mathcal{L} = \sqrt{\frac{1}{N} \sum_{i=1}^{N} ({\bf Y}_i - {\bf\hat{Y}}_i) ^ 2},
\end{equation}
which is a typical loss function for regression tasks.

\subsubsection{Modality Matching} 

Our goal in the modality matching task is to predict the probability that a pair of different modality data is actually from the same cell. Modality matching is very different from the modality prediction task in two regards: ({1}) it requires interactions between two modalities; and ({2}) it does not demand the model to give detailed predictions but it emphasizes pairings between two modalities. Therefore, we need to adjust the framework in graph construction and task-specific head correspondingly. 

Since two different modalities are presented as input in this task, we construct two cell-feature graphs. Cell-feature graph convolutions are then performed on these two graphs separately to obtain two embedding matrices $\mathbf{\hat{H}}_\text{m1} \in \mathbb{R}^{N \times d_1}$ and $\mathbf{\hat{H}}_\text{m2} \in \mathbb{R}^{N \times d_2}$. We set $d_1=d_2$ so they can be directly multiplied together. Thus, we calculate the cosine similarity between each cell pair as follows:
\begin{equation}
    \mathbf{S} = \mathbf{\hat{H'}}_\text{m1} \cdot \mathbf{\hat{H'}}_\text{m2}^T
\end{equation}
where $\mathbf{S} \in \mathbb{R}^{N \times N}$ denotes the symmetric score matrix; $\mathbf{\hat{H}}'_\text{m1}$ and $\mathbf{\hat{H}}'_\text{m2}$ indicate that we perform L2 normalization for each row (i.e., each cell) in $\mathbf{\hat{H}}_\text{m1}$ and $\mathbf{\hat{H}}_\text{m2}$ before we perform matrix multiplication. We further calculate the softmax function for each row and each column of ${\bf S}$ to convert scores to probabilities. Then we can express the loss function as follows:
\begin{align}
& \mathcal{L}_\text{match} = -\sum_{c_1=1}^N\sum_{c_2=1}^N\mathbf{Y}_{c_1,c_2}\log(\mathbf{P}^{r}_{c_1,c_2}) + \mathbf{Y}_{c_1,c_2}\log(\mathbf{P}^c_{c_1,c_2}) \\
& \text{with} \quad \mathbf{P}^r_{i,j} = \frac{e^{\mathbf{S}_{i,j}}}{\sum_{k=1}^N e^{\mathbf{S}_{i,k}}}, \quad 
\mathbf{P}^c_{i,j} = \frac{e^{\mathbf{S}_{i,j}}}{\sum_{k=1}^N  e^{\mathbf{S}_{k,j}}}, \nonumber 
\end{align}
where $\mathbf{Y} \in \mathbb{R}^{N \times N}$ denotes a binarized matching matrix that indicates the perfect matching (i.e., the ground truth label), and $\mathbf{P}_{i,j} \in \mathbb{R}^{N \times N}$ represents the probability that $i$-th data in modality 1 and $j$-th data in modality 2 are actually referring to the same cell.

In addition to the matching loss $\mathcal{L}_\text{match}$, we include a set of auxiliary losses to boost the performance, i.e., prediction losses and reconstruction losses: 
\begin{equation}
\begin{aligned}
    \mathcal{L}_\text{aux} &= \mathcal{L}_\text{pred12} + \mathcal{L}_\text{pred21} + \mathcal{L}_\text{recon11} + \mathcal{L}_\text{recon22}\\
    &= \frac{1}{N} \sum_{i=1}^{N} (\mathbf{X}_\text{m2} - f_{\theta_2}(\mathbf{\hat{H}}_\text{m1})) ^ 2 + \frac{1}{N} \sum_{i=1}^{N} (\mathbf{X}_\text{m1}  - f_{\theta_1}(\mathbf{\hat{H}}_\text{m2})) ^ 2\\
    &+ \frac{1}{N} \sum_{i=1}^{N} (\mathbf{X}_\text{m1}  - f_{\theta_1}(\mathbf{\hat{H}}_\text{m1})) ^ 2 + \frac{1}{N} \sum_{i=1}^{N} (\mathbf{X}_\text{m2}  - f_{\theta_2}(\mathbf{\hat{H}}_\text{m2})) ^ 2
\end{aligned}
\end{equation}
where $\mathbf{X}_\text{m1}$ and $\mathbf{X}_\text{m2}$ refer to the preprocessing results of two modalities respectively,  $f_{\theta_1}$ and $f_{\theta_2}$ each refers to a fully connected network with one hidden layer, which project the node embeddings $\mathbf{\hat{H}}$ to the particular target modality space. These auxiliary losses provide extra supervision for our model to encode the necessary information within the hidden space $\mathbf{\hat{H}}$.  Hence they have the potential to improve the robustness of the model and reduce the risk of overfitting. In the training phase, we jointly optimize $\mathcal{L}_\text{match}$ and $\mathcal{L}_\text{aux}$.

Lastly, in the inference phase, we introduce bipartite matching as an extra post-processing method to further augment our matching result. Specifically, we first use percentile threshold to filter the score matrix $\mathbf{S}$ in order to reduce the subsequent calculations, resulting in a symmetric sparse matrix $\mathbf{S}'$. We consider $\mathbf{S}'$ as an adjacency matrix of a bipartite graph, which helps us model the two different modality data and their inter-class relations. Thus, our goal (i.e., matching between two modalities) becomes a rectangular linear assignment problem, where we try to find a perfect matching that maximizes the sum of the weights of the edges included in the matching. We can effectively solve this problem through the Hungarian algorithm, also known as the Munkres or Kuhn-Munkres algorithm.

\subsubsection{Joint Embedding}
The target of the joint embedding task is to learn cell embeddings from multiple modalities and thus better describe cellular heterogeneity. Several complex metrics are enabled in this task, there often exist trade-offs between metrics and metrics (e.g. to remove the batch effect while retaining the batch information). To provide more training signals, we utilize both supervised and self-supervised losses to train our graph neural networks. Specifically, we first use the LSI for preprocessing to generate the input node features for two modalities and concatenate them as one joint modality, which allows us to construct a cell-feature graph. Based on the graph, we perform the proposed cell-feature graph convolution and generate the output cell node embedding in the same way as in Eq.~(\ref{eq:ro}). As suggested by the work~\cite{liu2021simultaneous}, cell type information plays a key role in the metrics of joint embedding evaluation and it is beneficial to extract $T$ dimensions from the hidden space to serve as supervision signals. Following this idea, we calculate the softmax function for $t \in \{1, ..., T\}$, $i \in \{1, ..., N\}$ with $N$ being the number of cells:
\begin{equation}
    \mathbf{\hat{Y}}_{i,t} = \frac{e^{\mathbf{\hat{H}}_{i,t}}}{\sum_{k=1}^T e^{\mathbf{\hat{H}}_{i,k}}}
\end{equation}
where  $\mathbf{\hat{Y}}$ is the probability that cell $i$ belongs to cell type $t$ and $T$ is set to be exactly equal to the total number of cell types. Then we introduce the loss functions: 
\begin{equation}
\begin{aligned}
    \mathcal{L} &= \mathcal{L}_\text{recon} + \mathcal{L}_\text{cell type} + \mathcal{L}_\text{regular}\\
    &= \frac{1}{N} \sum_{i=1}^{N} (\mathbf{X}_\text{LSI} - f_{\theta}(\mathbf{\hat{H}})) ^ 2
    + \sum_{t=1}^T\mathbf{Y}_t \log(\mathbf{\hat{Y}}_t) + \beta * \| \mathbf{\hat{H}}_{\mathcal{\tilde{J}}} \|_2
\end{aligned}
\end{equation}
where $f_{\theta}$ is a two-layer perceptron,  $\mathbf{Y} \in \mathbb{R} ^ {N \times T}$ is the sparse labels of cell types, and  $\mathbf{\hat{H}}_{\mathcal{\tilde{J}}}$ refers to the other hidden dimensions aside from the $T$ dimensions that have been exclusive to cell type information. Eventually, the output hidden space $\mathbf{\hat{H}}$ would contain cellular information required by the task, although the loss functions do not directly optimize the final metric.
\setlength{\textfloatsep}{1pt}
\section{Experiment}
In this section, we evaluate the effectiveness of our framework \method{} against three tasks and show how \method{} outperforms representative baselines over all three tasks by combining our general framework with task-specific design. Note that in this experiment, we follow the official settings and datasets in the multimodal single-cell data integration competition at NeurIPS 2021~\cite{luecken2021a} and we compare the performance of the proposed framework with that from the top winners in the competition. All source codes of our model have been integrated into the DANCE package~\cite{ding2022dance}.



\subsection{Modality Prediction}
\noindent\textbf{Datasets.} In the modality prediction dataset, each modality is provided with source data, preprocessed data (with default methods), and batch labels. Data statistics are shown in Table~\ref{dataset_task1} in Appendix~\ref{appendix:data}. Note that the GEX-ATAC and ATAC-GEX subtasks are not entirely symmetric, because in the GEX-ATAC task, the output dimension is deliberately reduced, where 10,000 ATAC features are randomly selected out of the original 116,490 dimensions.


\noindent\textbf{Settings and Parameters.} 
For our own model, we report the average performance of 10 runs. In each run, we reserved 15\% of training cells for validation and early stopping. In practice, several tricks help boost the performance: ({1}) utilizing initial residual instead of the skip connections in Eq.\ref{eq:v}, ({2}) using group normalization for aggregation results and dismissing the edge weight normalization stated in Eq.~\ref{eq:norm}. Besides, we empirically set our parameter $\alpha$ in Eq.~\ref{eq:a} to $0.5$. Additionally, in order to fit the high-dimensional ATAC features, we enabled node sampling.

\noindent\textbf{Baselines.}
In Table~\ref{task1lb}, we only show the teams that acquired top results in one or more subtasks in the competition, because they are officially required to declare their model details. For further comparison, we briefly detail each method: (1) {Baseline}, a truncated-SVD dimensionality reduction followed by linear regression.
(2) {Dengkw}, a well-designed model based on kernel ridge regression.
(3) {Novel}, an encoder-decoder structure with LSI preprocessing.
(4) {Living System Lab}, an ensemble model composed of random forest models, catboost\cite{prokhorenkova2018catboost} models, and k-nearest neighbors regression models.
(5) {Cajal}, a feed forward neural network with heavy feature selection guided by prior knowledge.
(6) {ScJoint}, an ensemble neural network model incorporated various strategies of preprocessing such as extracting TF-IDF features and filtering highly variable genes/features.

 
The source codes for all the methods above can be found in the
official github of the competition \footnote{\url{https://github.com/openproblems-bio/neurips2021_multimodal_topmethods}}. It can be seen that the existing models are relatively simple, mainly based on traditional machine learning algorithms and autoencoders. In contrast, our framework has a more advanced architecture, which provides the flexibility to different structural data (e.g. graph data) and different tasks. This makes our framework a very suitable backbone in the field of single-cell multimodal data integration.

\noindent\textbf{Results.}
As shown in Table~\ref{task1lb}, our method achieved the lowest overall loss in the competition (the lower, the better). An interesting observation is that there is no team lead consistently across all subtasks, resulting in individual s for each category, which is very different from the other two tasks in the competition. However, as far as we know, there is no team that worked only on one subtask. Such a phenomenon may be caused by three reasons: ({1}) the modality prediction task is the most competitive task in the competition, and many participating teams participated only in this task (including our own team). As a result, over 40 individual teams appeared on the final leaderboard. ({2}) the modality prediction task has only 1,000 cells in the private test dataset, therefore, certain variance exists in the evaluation results. ({3}) the diverse feature dimensionality and sparsity in different modalities raised additional challenges to the model's generalization ability. Compared to the other models, our GNN model presented consistently better performance across these four subtasks and became the overall winner in the competition.

Furthermore, we even improved our models after the competition, with modifications including: a learning-rate decay training strategy, more hidden units along with stronger regularization of dropout and weight decay.  Eventually, we've effectively strengthened our graph neural network model hence significantly improved our results, especially in the toughest subtask GEX-to-ADT, where the output for each cell is a 134-dimensional dense vector.  We now achieved an RMSE loss of $0.3809$ which is lower than the previous best score of $0.3854$ in the competition. Overall, the results prove the effectiveness of our GNN framework, and in some specific cases, \method{} has tremendous advantage in view of performance.  \vskip -1em

\begin{table}[t]
\footnotesize
\caption{RMSE for Modality Prediction (Task 1)${\downarrow{}{}}$. `*' indicates ensemble models.}
\vskip -1em
\begin{tabular}{lccccc}
\toprule
                                                                    & GEXADT                                                   & ADT2GEX                                                   & GEX2ATAC                                                  & ATAC2GEX                                                   & Overall \\ \midrule
Baseline                                                            & 0.4395                                                       & 0.3344                                                       & 0.1785                                                       & 0.2524                                                        & 0.3012  \\
Dengkw*                                                             & 0.3854                                                       & 0.3242                                                       & 0.1833                                                       & 0.2449                                                        & 0.2836  \\
Novel                                                               & 0.4153                                                       & $\bf 0.3177$                                                 & 0.1781                                                       & 0.2531                                                        & 0.2911  \\
LS. Lab*                                                            & 0.4065                                                       & 0.3228                                                       & $\bf 0.1774$                                                 & 0.2393                                                        & 0.2865  \\
Cajal                                                               & 0.4393                                                       & 0.3311                                                       & 0.1777                                                       & $\bf 0.2169$                                                  & 0.2891  \\
ScJoint*                                                            & 0.3954                                                       & 0.3247                                                       & 0.1785                                                       & 0.2377                                                        & 0.2840  \\ 
\method{}* & 0.3898                                                       & 0.3221                                                       & 0.1776                                                       & 0.2403                                                        & 0.2824  \\ \midrule
\begin{tabular}[c]{@{}l@{}}\method{} \\ (Single)\end{tabular}       & \begin{tabular}[c]{@{}l@{}}0.3885\end{tabular} & \begin{tabular}[c]{@{}c@{}}0.3242\end{tabular} & \begin{tabular}[c]{@{}c@{}}0.1778\end{tabular} & \begin{tabular}[c]{@{}c@{}}0.2315\end{tabular} & 0.2805  \\ \midrule
\begin{tabular}[c]{@{}l@{}}\method{}* \\ (Ensemble)\end{tabular}    & $\bf 0.3809$                                                 & 0.3223                                                       & 0.1777                                                       & 0.2310                                                        & $\bf 0.2780$  \\ \bottomrule
\end{tabular}
\label{task1lb}
\end{table}

\subsection{Modality Matching}
\noindent\textbf{Datasets.} The majority of the modality matching dataset is the same as the modality prediction dataset (as shown in Table~\ref{dataset_task2} in Appendix~\ref{appendix:data}, while several differences exist, including: ({1}) the number of testing cells; ({2}) the dimensionality of ATAC features; and ({3}) the inconsistent cell order among modalities. In the training data, samples' correspondence between different modalities is given, while for the test data, our goal is to find the correspondence.

\noindent\textbf{Settings and Parameters.} \label{setting:match}
The experimental settings are similar to the previous task, while some adjustments were made. For calculation convenience, we decoupled the propagation and transformation. Besides, batch labels are given in company with the test data, which provides a very strong prior knowledge for matching. We thus divided the test data into different batches, then matched samples that belong to the same batch. This trick dramatically reduced the search space, resulting in a significant performance boost. To be fair, we have confirmed that the winning solution also used the same strategy.

\noindent\textbf{Baselines.}
In Table~\ref{task2lb}, we only compare our models with the winning team and runner-up team in the competition. Next we briefly introduce those models:
(1) {Baseline}, first projecting one modality to the other with linear regression, and then searching for the nearest neighbors in the same modality.
(2) {GLUE~\cite{cao2022multi} (CLUE)}, the official winner, a variational auto-encoder (VAE) model supervised by three auxiliary losses.
(3) {Novel}, a feed-forward neural network directly supervised by matching loss.

\noindent\textbf{Results.}
As shown in Table~\ref{task2lb}, \method{} outperforms the winning team and the runner-up team with a very large margin. Note that we didn't create any models for this task during the competition since we focused on the modality prediction task. This is the reason why we don't have any results on the official leaderboard.  

The score of the metric can be roughly seen as the accuracy of predicting a right correspondence for each piece of data. Meanwhile the search space grows with the total number of cells in the test data. For example, in the test phase of the ADT-to-GEX subtask, we have 15,066 cells to match, thus for each piece of ADT data, we have 15,066 candidates in GEX data. Although we separated those cells into three batches, the rough expectation of the accuracy of randomly guessing is still as low as 1/5000, which can indicate the difficulty of this task. Thus, \method{} has already achieved very high performance (e.g. 0.08 in ADT-GEX subtask). Note that both team Novel's model and \method{} utilizes a symmetric matching algorithm, thus we have exactly the same performance for dual subtasks (e.g. GEX2ADT and ADT2GEX). Another interesting observation is that our proposed graph neural network model is especially good at GEX-ADT dual subtasks, where we improved the previous winning performance from $0.05$ to $0.08$.  \vskip -0.5em



\begin{table}[]
\caption{Performances for Modality Matching (Task 2)$\uparrow$.}
\vskip -1em
\footnotesize
\begin{tabular}{@{}lccccc@{}}
\toprule
                           & GEX2ADT                                                            & ADT2GEX                                                            & GEX2ATAC                                                           & ATAC2GEX                                                           & Overall         \\ \midrule
Baseline                   & 0.0000                                                                & 0.0000                                                                & 0.0001                                                                & 0.0001                                                                & 0.0001          \\
GLUE (CLUE)                       & 0.0495                                                                & 0.0516                                                                & 0.0560                                                                & 0.0583                                                                & 0.0539          \\
Novel                      & 0.0373                                                                & 0.0373                                                                & 0.0412                                                                & 0.0412                                                                & 0.0392          \\ \midrule
\method{} & \textbf{\begin{tabular}[c]{@{}l@{}}0.0810\end{tabular}} & \textbf{\begin{tabular}[c]{@{}c@{}}0.0810\end{tabular}} & \textbf{\begin{tabular}[c]{@{}c@{}}0.0630\end{tabular}} & \textbf{\begin{tabular}[c]{@{}c@{}}0.0630\end{tabular}} & \textbf{0.0720} \\ \bottomrule
\end{tabular}
\label{task2lb}
\vskip -1.5em
\end{table}

\subsection{Joint Embedding}
\noindent\textbf{Datasets.}
The training data of this task are basically the same as the modality prediction task. Moreover, data from different modalities are already aligned. Regarding the complementary data, there are two settings for this task. In the 'online' setting, the only data is features of two modalities. Meanwhile, in the 'pre-trained' setting, any external knowledge is acceptable. In this paper, we follow the second setting (i.e. pre-trained setting) and we only compare our results with other pre-trained models. Generally speaking, pre-trained models obtain better performance than online models. In our experiments, cell type labels are provided in the training phase, while test data consist of all the train cells and unseen test cells but are not along with cell type labels.

\noindent\textbf{Settings and Parameters.}
Considering that the input in this task is similar to modality matching, we followed settings in Section~\ref{setting:match}. 

\noindent\textbf{Baselines.}
We briefly describe the other three models in Table~\ref{task3lb}:
(1) {Baseline}, a concatenation of PCA results of two modalities.
(2) {Amateur (JAE)}, the official winner, an auto-encoder model that incorporates extra supervision (cell annotations). The model is adapted from scDEC~\cite{liu2021simultaneous}.
(3) {GLUE~\cite{cao2022multi}}, an auto-encoder model guided by an external knowledge graph. 

\noindent\textbf{Results.}
As shown in Table~\ref{task3lb}, our \method{} significantly outperforms the other two models in GEX-ADT joint embedding task, with an improvement over 0.1 according to the average metric.

\begin{table*}[t]
\vskip -1em
\small
\caption{Performances for GEX-ADT Joint Embedding (Task 3)$\uparrow$.}
\vskip -1em
\begin{tabular}{@{}lccccccc@{}}
\cmidrule(r){1-8}
             & NMI cluster/label                      & Cell type ASW           & Cc\_con         & Traj\_con       & Batch ASW     & Graph connectivity           & Average metric           \\ \cmidrule(r){1-8}
Baseline     & 0.6408                   & 0.5266                   & \textbf{0.9270} & 0.8325          & 0.7982          & 0.8945                   & 0.7699                   \\
Amateur (JAE) & 0.7608                   & 0.6043                   & 0.7817          & \textbf{0.8631} & 0.8432          & 0.9700                   & 0.8039                   \\
GLUE         & 0.8022                   & 0.5759                   & 0.6058          & 0.8591          & \textbf{0.8800} & 0.9506                   & 0.7789                   \\ \cmidrule(r){1-8}
\method{}    & \textbf{0.8499} & \textbf{0.6496} & 0.7084 & 0.8532 & 0.8691 & \textbf{0.9708} & \textbf{0.8168} \\ \cmidrule(r){1-8}
\end{tabular}
\label{task3lb}
\vskip -2em
\end{table*}

\begin{figure}[t]%
     \centering
     \label{fig:x}{\includegraphics[width=0.7\linewidth]{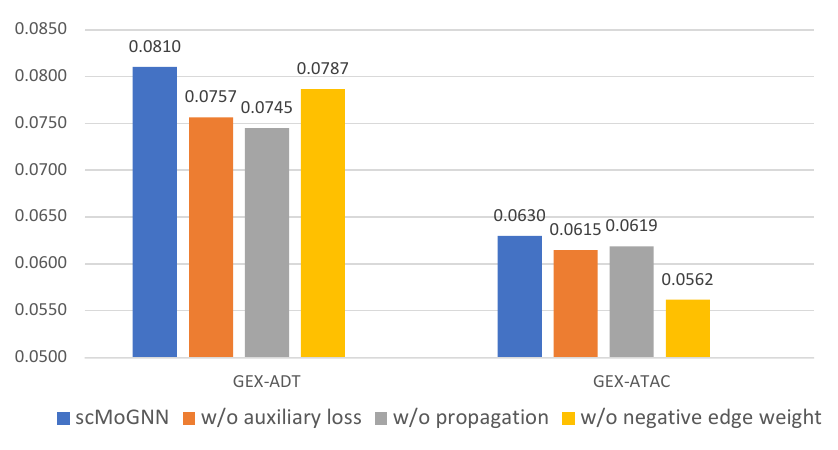} }%
    \qquad
\vskip -1em
\caption{Ablation study for the modality matching task.}
\label{fig:abl}%
\end{figure}

\begin{figure}[t]%
     \centering
     \label{fig:x}{\includegraphics[width=0.7\linewidth]{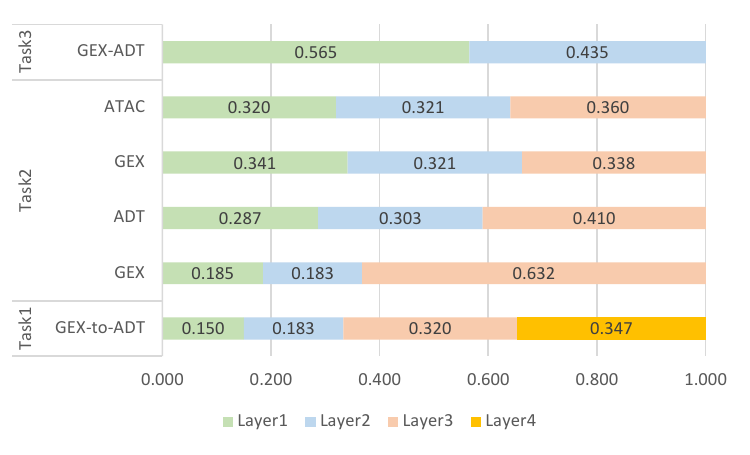} }%
    \qquad
\vskip -1em
\caption{Parameter analysis of layer weight $\bf w$.}
\label{fig:prm}%
\end{figure}

\subsection{Ablation Study}
Throughout the previous sections, we have examined that our graph neural network general framework is suitable for all these tasks in single-cell multimodal data integration. In this subsection, we investigate if we really benefit from graph structural information. We take the modality matching task as an example. In the modality matching task we use decoupled GNNs, thus, we can easily remove the graph structural information by eliminating the propagation layers. The result is referred to as ``w/o propagation'' in Figure~\ref{fig:abl}. The performance significantly drops from $0.0810$ to $0.0745$ in the GEX-ADT subtask and from $0.0630$ to $0.0562$ in the GEX-ATAC subtask, respectively. These observations indicate that the graph structural information extracted by the propagation layers indeed helped the performance of our method significantly. We also examined the importance of our auxiliary loss, shown in Figure~\ref{fig:abl}. Without the supervision of auxiliary losses, \method{} lost a lot of generalization ability, behaving as poorly as without graph structural information. \vspace{-2em} 

\subsection{Parameter Analysis}
We analyzed an important parameter in our framework, i.e. $\bf w$ in Eq.~\ref{eq:ro}, in order to gain a deeper insight on \method{}. Specifically, $\bf w$ is a learnable parameter that controls the weight between each propagation layer. Intuitively, the value of $\bf w$ can prove to us the effectiveness of graph structural information and help us understand how much higher-order structural information is valued by models in different tasks. Therefore we show values of $\bf w$ learned by \method{} in different tasks in Figure~\ref{fig:prm}. Note that in different tasks we have different numbers of layers, in modality prediction  we have 4 layers and in modality matching and joint embedding we have 3 layers and 2 layers, respectively. The results consistently show that \method{} tends to synthesize the information in each layer, not just limited to the shallow layer, which suggests that the information of the higher-order graph structure is indeed effective. As for more details, joint embedding depends more on local information, which exists in source input data. While in modality prediction, more higher-order information is referenced, indicating that the model needs to enrich more information from similar cells or similar features. This can be explained from the need for more detailed information in modality prediction. 





\section{Conclusion}
In this work, we proposed a general framework \method{} based on GNNs for multimodal single-cell data integration. It can be broadly applied on all three key tasks, modality prediction, modality matching and joint embedding, from the NeurIPS 2021 Competition. 
Our framework \method{} is able to capture high-order structural information between cells and features. To the best of our knowledge, we are the first to apply GNNs in this field.
Our method officially wins first place in the overall ranking of the modality prediction task and now outperforms all models from three tasks on the leaderboard with remarkable advantage.

\begin{acks}
This research is supported by the National Science Foundation (NSF) under grant numbers IIS1714741, CNS1815636, IIS1845081, IIS1907704, IIS1928278, IIS1955285, IOS2107215 and IOS2035472, the Army Research Office (ARO) under grant number W911NF-21-1-0198, the Home Depot, Amazon, Cisco Systems Inc and SNAP.
\end{acks}


\bibliographystyle{ACM-Reference-Format}
\bibliography{sample}

\newpage

\appendix
\section{Details of Metrics in Task 3}
\label{sec:metric_task3}

\subsection{Biology Conservation Metrics}
\begin{itemize}
    \item \textbf{NMI cluster/label:} Normalized mutual information (NMI) compares the overlap of two clusterings. We use NMI to compare the cell type labels with an automated clustering computed on the integrated dataset (based on \texttt{Louvain clustering} \footnote{\url{https://en.wikipedia.org/wiki/Louvain_method}}). NMI scores of 0 or 1 correspond to uncorrelated clustering or a perfect match, respectively. Automated Louvain clustering is performed at resolution ranges from 0.1 to 2 in steps of 0.1, and the clustering output with the highest NMI with the label set is used.
    
    \item \textbf{Cell type ASW:} The silhouette width metric indicates the degree to which observations with identical labels are compact. The average silhouette width (ASW)~\cite{batool2021clustering}, which ranges between -1 and 1, is calculated by averaging the silhouette widths of all cells in a set. We employ ASW to determine the compactness of the resulting embedding's cell types. The ASW of the cluster is calculated using the cell identity labels and scaled to a value between 0 and 1 using the equation:
    \begin{equation}
     {ASW} = ( ASW_C + 1 ) / 2
    \end{equation}
    where C denotes the set of all cell identity labels.

    \item \textbf{Cell cycle conservation:} The cell cycle conservation score serves as a proxy for the preservation of the signal associated with gene programs during data integration. It determines the amount of variance explained by cell cycle per batch prior to and following integration. The differences in variance before ${Var_{before}}$ and variance after ${Var_{after}}$ are aggregated into a final score between 0 and 1, using the equation:
    \begin{equation}
     {CC_{conservation}} = 1 - \frac{|{Var_{after}} - {Var_{before}}|}{{Var_{before}}}
    \end{equation}
    where values near to 0 suggest less conservation of variance explained by the cell cycle, while 1 represents complete conservation.

    \item \textbf{Trajectory conservation:} The conservation score of a trajectory is a proxy for the conservation of a continuous biological signal within a joint embedding. We compare trajectories computed after integration for relevant cell types that depict a continuous cellular differentiation process to trajectories computed per batch and modality using this metric. 
    The conservation of the trajectory is quantified via Spearman’s rank correlation coefficient~\cite{sedgwick2014spearman}, $s$, between the pseudotime values before and after integration. The final score is scaled to a value between 0 and 1 using the equation:
     \begin{equation}
     {trajectory conservation = (s + 1) / 2}
    \end{equation}
    where a value of 1 or 0 indicates that the cells on the trajectory are in the same order before and after integration, or in the reverse order.
\end{itemize}

\subsection{Batch Removal Metrics}
\begin{itemize}
    \item \textbf{Batch ASW:} The ASW is used to quantify batch mixing by taking into account the incompatibility of batch labels per cell type cluster. We consider the absolute silhouette width, on batch labels per cell, in particular. Here, zero shows that batches are thoroughly mixed, but any variation from zero indicates the presence of a batch effect. We rescale this score so that higher values imply better batch mixing and use the equation below to determine the per-cell type label, j:

    \begin{equation}
    \operatorname{batch} A S W_{j}=\frac{1}{\left|C_{j}\right|} \sum_{i \in C_{j}} 1-|s(i)|
    \end{equation}
    where $C_j$ is the set of cells with the cell label j and $|C_j|$ denotes the number of cells in that set. 
    To obtain the final $batchASW$ score, the label-specific $batchASW_{j}$ scores are averaged:
    
    \begin{equation}
    \text { batch } A S W=\frac{1}{|M|} \sum_{j \in M} \text { batch } A S W_{j}
    \end{equation}
    where M is the set of unique cell labels. A $batchASW$  value of 1 suggests optimal batch mixing, whereas a value of 0 indicates severely separated batches.
    
    \item \textbf{Graph connectivity:} The graph connectivity metric determines whether cells of the same kind from various batches are embedded close to one another. This is determined by computing a k-nearest neighbor (kNN) graph using Euclidean distances on the embedding. Then, we determine if all cells with the same cell identity label are connected in this kNN graph. For each cell identity label $c$, we generate the subset kNN graph $G=(N_c; E_c)$, which contains only cells from a given label.
    Using these subset kNN graphs, we compute the graph connectivity score:
    \begin{equation}
    g c=\frac{1}{|C|} \sum_{c \in C} \frac{\left|L C C\left(G\left(N_{c} ; E_{c}\right)\right)\right|}{\left|N_{c}\right|}
    \end{equation}
    where $C$ represents the set of cell identity labels, 
    $LCC()$ denotes the number of nodes in the largest connected component of the graph, and $N_c$ is the number of nodes with cell identity $c$.
    The resulting score ranges from 0 to 1, where 1 means that all cells with the same cell identity are connected in the integrated kNN graph, while 0 indicates that no cell is connected in the network.
\end{itemize}

\subsection{Metric Aggregation}
Due to the differing nature of each metric, each metric would be assigned with a weight. An overall weighted average of batch correction and bio-conservation scores will be computed via the equation:
\begin{equation}
    S_{\text {overall }, i}=0.6 \cdot S_{b i o, i}+0.4 \cdot S_{b a t c h, i}
\end{equation}

\section{Data Statistics}
\label{appendix:data}
The Table~\ref{dataset_task1} and Table \ref{dataset_task2} provide statistics about dataset used in modality prediction task and modality matching task respectively.

\begin{table}[tb]
\vskip -1em
\small
\caption{Dataset Statistics of modality prediction task. The number of feature dimensions, train/test samples, and batches.}
\vskip -1em
\begin{tabular}{c|cccc}
\toprule
 & \textbf{GEX-ADT} & \textbf{ADT-GEX} & \textbf{GEX-ATAC} & \textbf{ATAC-GEX} \\ \midrule
Source Dim & 13,953 & 134 & 13,431 & 116,490 \\
Target Dim & 134 & 13,953 & 10,000 & 13,431 \\
Train Cells & 66,175 & 66,175 & 42,492 & 42,492 \\
Test Cells & 1,000 & 1,000 & 1,000 & 1,000 \\
Train Batches & 9 & 9 & 10 & 10 \\
Test Batches & 3 & 3 & 3 & 3 \\ \bottomrule
\end{tabular}
\label{dataset_task1}
\end{table}

\begin{table}[tb]
\vskip -1em
\small
\caption{Dataset Statistics of modality matching task. The number of feature dimensions, train/test samples, and batches.}
\vskip -1em
\begin{tabular}{c|cccc}
\toprule
 & \textbf{GEX-ADT} & \textbf{ADT-GEX} & \textbf{GEX-ATAC} & \textbf{ATAC-GEX} \\ \midrule
Source Dim & 13,953 & 134 & 13,431 & 116,490 \\
Target Dim & 134 & 13,953 & 116,490 & 13,431 \\
Train Cells & 66,175 & 66,175 & 42,492 & 42,492 \\
Test Cells & 15,066 & 15,066 & 20,009 & 20,009 \\
Train Batches & 10 & 10 & 10 & 10 \\
Test Batches & 3 & 3 & 3 & 3 \\ \bottomrule
\end{tabular}
\vspace{0.5em}
\label{dataset_task2}
\end{table}

\section{Reproducibility}

\subsection{source code}
All the source code of winning solutions can be found at official github (\url{https://github.com/openproblems-bio/neurips2021_multimodal_topmethods}). These codes have been officially verified thus reproducibility is ensured.

For the new models we developed after the competitions, all source codes have been integrated into the DANCE package~\cite{ding2022dance}.


\end{document}